\documentclass[conference]{IEEEtran}
\usepackage[utf8]{inputenc}
\usepackage[pdftex]{graphicx}
\usepackage{soul}
\usepackage{epstopdf}
\usepackage{multirow}
\usepackage{amsmath}
\usepackage{mathtools}
\usepackage{authblk}
\usepackage{cite}
\usepackage{cleveref}
\usepackage{latexsym}
\usepackage[justification=centering]{caption}
\usepackage{amsthm}
\usepackage{balance}
\usepackage{relsize}
\usepackage{color}
\usepackage{amssymb}
\usepackage{eucal}
\usepackage{url}
\usepackage{comment}
\usepackage{booktabs,subcaption,amsfonts,dcolumn}
\usepackage[nodisplayskipstretch]{setspace}
\usepackage{lipsum}
\usepackage{acronym}
\usepackage{makecell}

\newcolumntype{L}{>{\centering\arraybackslash}m{6cm}}

\newacro{kpi}[KPI]{key performance indicator}
\newacro{ai}[AI]{artificial intelligence}
\newacro{fm}[FM]{foundation model}
\newacro{llm}[LLM]{large language model}
\newacro{tdoc}[Tdoc]{technical documents}
\newacro{cr}[CR]{change request}
\newacro{ml}[ML]{machine learning}
\newacro{bert}[BERT]{bidirectional encoder representations from transformers}
\newacro{gpt}[GPT]{generative pre-trained transformers}
\newacro{rf}[RF]{radio frequency}
\newacro{ran}[RAN]{radio access network}
\newacro{sa}[SA]{system architecture}
\newacro{ct}[CT]{core network and terminals}
\newacro{qos}[QoS]{quality of service}
\newacro{3gpp}[3GPP]{3rd Generation Partnership Project}
\newacro{wg}[WG]{working group}
\newacro{5ga}[5G-A]{5G Advanced}
\newacro{bpe}[BPE]{byte pair encoding}

\title{Understanding Telecom Language Through Large Language Models}
\author{Lina~Bariah, Hang~Zou, Qiyang~Zhao, Belkacem~Mouhouche, Faouzi~Bader, and Merouane~Debbah}
        
\affil{Technology Innovation Institute, 9639 Masdar City, Abu Dhabi, UAE \\ Email: firstname.lastname@tii.ae} 

\begin{document}

\maketitle

\begin{abstract}
The recent progress of artificial intelligence (AI) opens up new frontiers in the possibility of automating many tasks involved in Telecom networks design, implementation, and deployment. This has been further pushed forward with the evolution of generative artificial intelligence (AI), including the emergence of large language models (LLMs), which is believed to be the cornerstone toward realizing self-governed, interactive AI agents. Motivated by this, in this paper, we aim to adapt the paradigm of LLMs to the Telecom domain. In particular, we fine-tune several LLMs including BERT, distilled BERT, RoBERTa and GPT-2, to the Telecom domain languages, and demonstrate a use case for identifying the \ac{3gpp} standard working groups. We consider training the selected models on \ac{3gpp} \ac{tdoc} pertinent to years 2009-2019 and predict the \ac{tdoc} categories in years 2020-2023. The results demonstrate that fine-tuning BERT and RoBERTa model achieves 84.6\% accuracy, while GPT-2 model achieves 83\% in identifying 3GPP working groups. The distilled BERT model with around 50\% less parameters achieves similar performance as others. This corroborates that fine-tuning pretrained LLM can effectively identify the categories of Telecom language. The developed framework shows a stepping stone towards realizing intent-driven and self-evolving wireless networks from Telecom languages, and paves the way for the implementation of generative AI in the Telecom domain. 
\end{abstract}

\begin{IEEEkeywords}
Generative AI, Large Language Models, Pre-trained Transformer, Telecom Language, 3GPP
\end{IEEEkeywords}
\begin{table*}
\centering
\caption{BERT fine-tuning: Domains and Data} \label{tab:domain}
\begin{tabular}{ | c | c | L | } 
  \hline \hline
  Ref. & Domain & Data \\
  \hline \hline
   \cite{beltagy2019scibert} & General Science  & Scientific publication \\ 
  \hline
  \cite{lee2020patent} & Multi-domain Patents & Google Patents Public dataset \\ 
  \hline
  \cite{puri2019zero} & General domain Q\&A & Social commenting platforms \\ 
  \hline
  \cite{howard2018universal} & General domain & Wikipedia articles \\ 
  \hline
  \cite{myagmar2019cross} & Cross-domain sentiment analysis & Amazon reviews dataset \\ 
  \hline
  \cite{prabhu2021multi} & Customers delivery and pickup services (open-domain) & AG’s News Corpus \& TREC dataset \\ 
  \hline
  \cite{fabien2020bertaa} & Authorship attribute & Public datasets, e.g., IMDb \\ 
  \hline
  \cite{chen2022long} & Chinese language & Chinese web news \\ 
  \hline
   \cite{lagutina2022topical} & Russian language & OpenCorpora (Russian online media and Russian Wikipedia) \\ 
  \hline
  \cite{antoun2020arabert} & Arabic language & Arabic media \& news (Modern Standard Arabic (MSA) \& Dialectal Arabic (DA)) \\ 
  \hline
  \cite{he2020infusing} & Healthcare & Wikipedia articles \\ 
  \hline
  \cite{holm2021bidirectional} & Telecom Q\&A & TeleQuAD \\ 
  \hline\hline
  \textit{Our work} & \textit{Telecom Language Understanding} & \textit{3GPP technical documents and specifications} \\  
  \hline\hline
\end{tabular}
\end{table*}
\section{Introduction}
In the last couple of decades, considerable efforts have been devoted to push the frontiers of wireless technologies in order to achieve \acp{kpi} pertinent to latency, reliability, spectral and energy efficiencies, to name a few, through the exploitation of \ac{ai} as a network orchestrator. Recently, parallel initiatives have been focused on advancing the paradigm of self-evolving networks (under several names including autonomous networks, zero-touch networks, self-optimizing/configuring/healing networks, etc.), through the evolution of native intelligent network architecture \cite{wang2023road}. However, recent developments are revolving around realizing adaptivity, in which wireless networks functionalities can be autonomously adjusted to fit within a particular scenario. The ultimate vision of self-evolving networks goes way beyond adaptivity and automation. In particular, it expands toward realizing perpetual sustainability of network performance and the flexibility to accommodate highly complex, and sometimes unfamiliar, network scenarios, and hence, this necessitates generalized, inclusive, and multi-functional schemes that are capable of handling diverse network conditions. 
Accordingly, conventional \ac{ai} algorithms are highly probable to fall behind in fulfilling the required performance, and therefore, a radical departure to more innovative \ac{ai}-driven approaches is anticipated to shape the future of next generation wireless networks. 

\Acp{fm} was coined by Stanford Center for Research on Foundation Models (CRFM) in 2021 and have attracted a considerable attention as generalized models that are capable of handling a wide range of downstream tasks \cite{bommasani2021opportunities}. In particular, \acp{fm} are extremely large neural networks that are trained over massive unlabeled datasets, in a self-supervised fashion, allowing several opportunities to be reaped with reduced time and cost (that would be unbearable in case of human labeling). Rapidly after being developed, \acp{fm} have found their applications in several domains, including text classification and summarizing, sentiment analysis, information extraction, and image captioning. While \acp{fm} were not aimed to follow a particular model or application, language-related models, i.e., \acp{llm}, are currently one of the most common subfield of \acp{fm}, which rely on the principle of pretraining large models over a large-scale corpus. Such pretrained large models, e.g., \ac{bert} \cite{devlin2018bert} and \ac{gpt} \cite{radford2019language}, can be further fine-tuned in various downstream tasks, and hence, avoid the cost of retraining large models from scratch in the new domains. 

\subsection{Related Work}
Focusing on text generation-related tasks, language models trained on large corpus can successfully understand the natural language, and create human-like language responses according to the specific tasks. Several domain-specific variations of well-known pretrained language models were presented in the literature to demonstrate the opportunities that can be obtained from domain-specific fine-tuning and retraining. In \cite{beltagy2019scibert}, the authors proposed SCIBERT, a BERT-based language model that is fine-tuned to the scientific domain, where it was trained over corpus from scientific publications. The authors in \cite{lee2020patent} have considered fine-tuning BERT model using Google Patents Public datasets to perform patents classification. Furthermore, generative language model, based on multiple choice question answering, is fine-tuned using social commenting platforms in \cite{puri2019zero}, in order to realize zero-shot text classification. From a different perspective, the authors in \cite{howard2018universal} proposed a Universal Language Model Fine-tuning (ULMFiT) approach for fine-tuning generative large models for enhanced text classification. The proposed scheme in \cite{howard2018universal} demonstrated reduced error by 18-24\% up to six-class text classification, when tested over general Wikipedia articles. Cross-domain sentiment analysis through fine-tuning BERT and XLNet models is proposed in \cite{myagmar2019cross}, in which the fine-tuned model showed promising results with less amount of data. The authors in \cite{prabhu2021multi} explored several active learning strategies to adapt \ac{bert} model into customers transactions application to classify transactions to different market-related categories, for improved market demands understanding. Targeting different domain, the work in \cite{fabien2020bertaa} presents BertAA, a framework for \ac{bert} fine-tuning for authorship classification purposes, in which public datasets, e.g., IMDb, are utilized to refine the \ac{bert} model and enable it to extract the characteristics of authors' identities from the provided text. The proposed work in \cite{fabien2020bertaa} showed 5.3\% improved in the authorship attribute task. From a language perspective, multi-lingual and single-lingual frameworks were presented in the literature to fine-tune/retrain a pre-trained \ac{bert} model in order to allow the model to deal with different languages, e.g. Chinese \cite{chen2022long}, Russian \cite{lagutina2022topical}, Arabic \cite{antoun2020arabert}. The presented results in \cite{chen2022long}-\cite{antoun2020arabert} demonstrated the robustness of \ac{bert} as a large model for different languages. For healthcare applications, the authors in \cite{he2020infusing} provided a framework for disease name recognition, where a \ac{bert} model, fine-tuned using data pertinent to disease knowledge, demonstrated an enhanced performance compared to the literature.

\subsection{Contributions}
While the field of domain-specific fine-tuning of large generative models is very active and several contributions were presented for different domains, the telecom domain is still almost untouched. We strongly believe that adapting various large generative models to the telecom domain is a key building block in the development of self-evolving networks, where such models can play an essential role through the different stages of designing, building, and operating wireless networks. The advantages of large Telecom language models are envisioned to be particularly important with the rise of generative agents paradigm \cite{park2023generative}, in which \ac{llm} implemented at Telecom networks will require a comprehensive understanding of the Telecom terminologies, and their relationship with different network operational and configuration functions, in order to enable them to communicate meaningfully and to perform Telecom-specific downstream tasks when implemented in future wireless networks. 

Within this context, in \cite{holm2021bidirectional}, the authors have focused on adapting BERT-like model to the telecom domain, where the considered model is pretrained/fine-tuned in order to perform a question answering downstream task within the telecom domain. Note that the work in \cite{holm2021bidirectional} is constrained by the small dataset used (few hundreds of technical documents and web articles), which was prepared in a manual manner as follows. The data were acquired from technology specification files of the \ac{3gpp}, and it was collected from 347 telecom-related documents, resulting in 2,021 question-answer pairs only. It is worthy to note that the dataset used in \cite{holm2021bidirectional} is not publicly available. For enabling a holistic understanding of Telecom language, a comprehensive dataset comprising a wide-range of technical discussion pertinent to different network operational, configuration, and design parameters need to be generated and used in the pretraining/fine-tuning process. Motivated by this, in this paper, we develop a framework for adapting pretrained generative models, including \ac{bert}, DistilBERT, RoBERTa, and GPT-2 models, to the Telecom domain, through exploiting a huge number of technical documents that consist of technical specification from \ac{3gpp} standard. Among different language models, the selection of considered models are motivated by the fact that it generates a contextual representation for each word, while considering previous and following words, rendering it a well-suited model for technical text classification.

The main contributions of our work are summarized as follows:
\begin{enumerate}
\item Create an annotated large Telecom datasets from \ac{3gpp} technical specification of various \ac{wg}, including technical pertinent to \ac{rf} spectrum usage, network architecture, radio interface protocols, signaling procedures, and mobility management,  network architecture, system interfaces, security, \ac{qos}, network management, routing, switching, and control functions.
\item Adapt the pre-trained BERT, DistilBERT, RoBERTa, and GPT-2 model into the Telecom domain, through fine-tuning the models for \ac{3gpp} \acp{tdoc} text classification. The fine-tuned models allow to identify a particular technical text, related to \ac{3gpp} cellular architecture category, i.e., \ac{ran}, \ac{sa}, or \ac{ct}, with characterizing the \ac{wg} corresponding to each category.
\end{enumerate}

The remaining of the paper is organized as follows. In Sec. \ref{method} we detail the developed approach to adapt the pre-trained models to Telecom domain, including solutions on data collection, data pre-processing, and model fine-tuning. Experimental results with performance analysis are discussed in Sec. \ref{results}. Finally, the paper is concluded in Sec. \ref{conclusion}.

\section{Method}
\label{method}

\subsection{\acp{llm} for Telecom Language Classification}
In this work, we use \ac{bert}, DistilBERT, RoBERTa, GPT-2 language models, which are trained on large amounts of unlabeled textual data using self-supervised or contrastive learning \cite{devlin2018bert}. These models can be adapted to various downstream tasks via fine-tuning. Specifically, the architecture of BERT and its variants allows it to understand the context and meaning of words in a sentence by taking into account the surrounding words on both sides of the target word. This bidirectional approach helps the pre-trained model to capture more complex relationships between words and their contextual meaning, making it a powerful tool for text classification. 

The following models are implemented in our work: 1) Pretrained BERT-Base (uncased): contains 12 layers, 768 hidden units, 12 self-attention heads, and 110M parameters; 2) DistilBERT: a lighter version of BERT-Base (uncased) with 40\% less parameters, which is particularly useful for wireless network with constrained resources; 3) RoBERTa: contains the same architecture as BERT, with byte-level \ac{bpe} as a tokenizer is used, which operates at the byte level instead of the traditional character or subword levels; 4) GPT-2: the smallest version with 6 layers, 36 hidden units, 48 self-attention heads, and 124M parameters. A linear classification layer with SoftMax function is added to the pre-trained models to produce the \ac{wg}s.  

In order to adapt the selected models into the desired Telecom domain for the downstream task of text classification, we consider the single-task single-label fine-tuning approach \cite{sun2019fine} for \ac{3gpp} \ac{tdoc} classification. A cross entropy loss function is used to update the pre-trained model weights. For efficient fine-tuning, we employed a batch size of 32, ensuring a balance between computational efficiency and memory requirements. The learning rate was set to 2e-5, enabling gradual convergence to an optimal solution. To prevent overfitting, we applied L2 regularization with a rate of 0.01. Also, F1 score is considered to evaluate the performance of the tuned models. 

\subsection{3GPP Technical Document Dataset}
\ac{3gpp} is the main Standard Developing Organization (SDO) in the area of Telecommunication. The universal standards for 3G, 4G and 5G have been developed by \ac{3gpp} since 1999. \ac{3gpp} works with \ac{tdoc} contributed by companies during the development phase and produces technical specifications as a final output. The specification work is carried out in Technical Specification Groups (TSGs). There are three Technical Specifications Groups: \ac{ran}, \ac{sa}, and \ac{ct}. Each TSG consists of multiple \ac{wg} focused on specific areas, ranging from radio access network specifications, core network specifications, service requirements and specifications, and architecture and protocols for mobile communication systems, to \ac{qos} and performance requirements, security and privacy in mobile communication systems, interoperability and compatibility requirements, network management and operation, and testing and certification procedures. These topics are further divided into specific subtopics, and each \ac{tdoc} file may focus on one or more of these areas. The content of \ac{tdoc} files is typically technical and detailed, intended for experts and engineers involved in the development and implementation of mobile communication systems. Thus, The ability to classify a text into one of the \ac{wg} requires a deep understanding of the functions and scope of each group.

In this paper, the technical documents are acquired from the 3GPP website. The collected files belong to years 2009-2023 and include technical specifications put by different \acp{wg}, including, RAN1, RAN2, RAN3, RAN4, RAN5, SA1, SA2, SA3, SA4, SA5, SA6, CT1, CT3, CT4, CT6. The \ac{tdoc} files are available as ZIP files, and accordingly Apache Tika application \cite{AT} is used to unzip and extract the information from the files. Table \ref{tab:size} demonstrates the size of the dataset acquired from \ac{3gpp} \acp{wg}, where documents belonging to years 2009-2019 are used for training, while documents related to years 2020-2023 are exploited for testing. 

\begin{figure*}[h!]
\centering
  \includegraphics[width=0.9\textwidth]{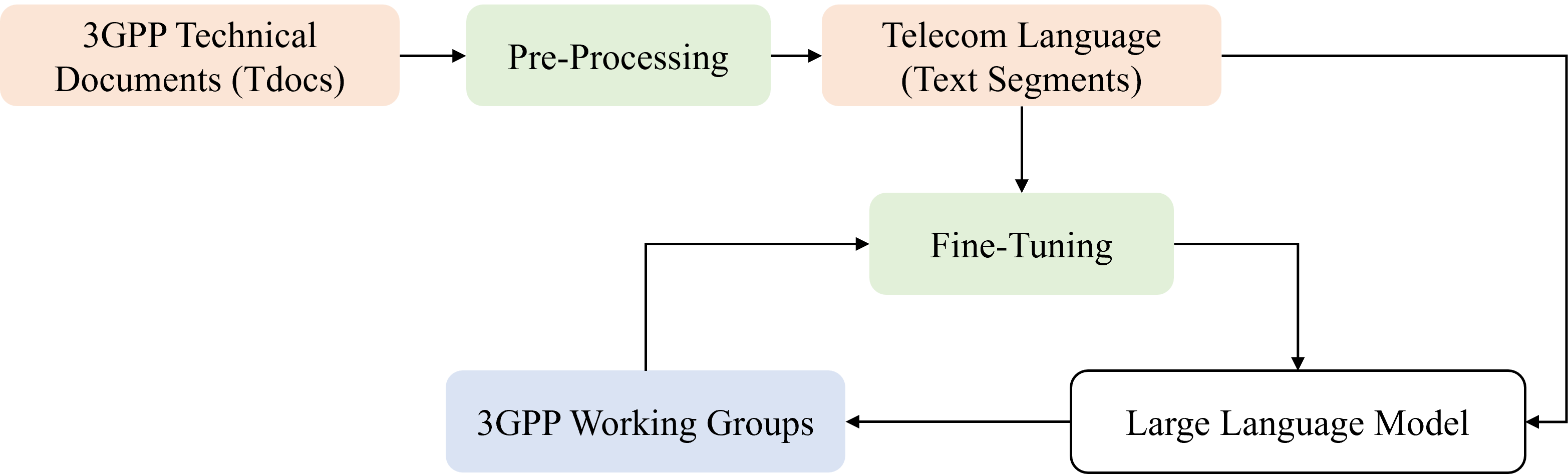}
  \caption{Pipeline for Adapting LLM to 3GPP Technical Language}
  \label{fig:Method}
\end{figure*}

\begin{table}
\centering
\caption{3GPP \ac{tdoc} files in different \ac{wg}s and years} \label{tab:size}
\begin{tabular}{ | c | c | c | c | } 
  \hline \hline
  \makecell{\ac{wg}} & \makecell{Train\\ ('09-'19)} & \makecell{Train \\('15-'19)} & \makecell{Test \\ ('20-'23)}\\
  \hline 
  RAN1 & 83905 & 56969 & 39230 \\ 
  \hline 
  RAN2 & 82853 & 55189 & 38744 \\ 
  \hline 
  RAN3 & 36651 & 22981 & 20219\\ 
  \hline 
  RAN4 & 43845  & 28953 & 43482 \\ 
  \hline
  RAN5 & 50812 & 31706 & 32456\\ 
  \hline
  SA1 &19497  & 11282  & 10086 \\ 
  \hline
  SA2 & 64065 & 42931 & 43860 \\ 
  \hline
  SA3 & 19546  & 13903  & 13815 \\ 
  \hline
  SA4 &6583  & 1776  & 5128  \\ 
  \hline
  SA5 & 28044  & 15031  & 13040  \\ 
  \hline
  SA6 & 9010  & 9010  & 10360\\ 
  \hline
  CT1 & 30990 & 20840 & 18910\\ 
  \hline
  CT3 & 22269 & 12734  & 12584\\ 
  \hline
  CT4 & 28245  & 14731 &  13109\\ 
  \hline
  CT6 & 4446  & 2213  & 1571\\ 
  \hline
  \hline
  Total & 520761  & 340244  & 316594 \\ 
  \hline\hline
\end{tabular}
\end{table}

\subsection{Data Pre-Processing}
We pre-process the 3GPP \ac{tdoc} files via following steps:
\begin{enumerate}
    \item Parse the HTML tags in the text and return the text content without any HTML tags using \textit{BeautifulSoup}.
    \item Remove any URLs (web links) from the text: identify the regex pattern that matches URLs starting with either "http" or "https" and may include alphanumeric characters, special characters, and encoded characters.
    \item Remove tables from the parsed HTML document using \textit{BeautifulSoup}. 
    \item Divide each document into multiple text segments with different number of words extracted from natural language toolkit (NLTK). This allows us to evaluate the model's capability of understanding technical descriptions in different lengths. 
    \item Remove headers, footers, captions, and pseudo codes, while ensuring each paragraph doesn't exceed a particular maximum length. Also, we eliminate the references section and all the text afterward. 
    \item Remove \acp{cr}, drafts, templates due to their limited technical information.    
\end{enumerate}


\section{Experiment Results and Discussions}
\label{results}
In this section, we present experimental results to demonstrate the accuracy of the fine-tuned \acp{llm} in understanding and classifying technical text within the Telecom domain. We split \ac{3gpp} \ac{tdoc} into training, validation, and test datasets. Specifically, the test set contains textual segments of \ac{tdoc}s from 2020 to 2023 (April). The training and validation sets contains: 1) \ac{tdoc}s from 2010 to 2019 ('10-'19); and 2) \ac{tdoc}s from 2015 to 2019 ('15-'19). The proportion of these two datasets is 80\% and 20\%. The number of words within a textual segment in training, validation and test set is $200$ in what follows without further mentioning.

\begin{figure}[h!]
\centering
\includegraphics[width=1\linewidth]{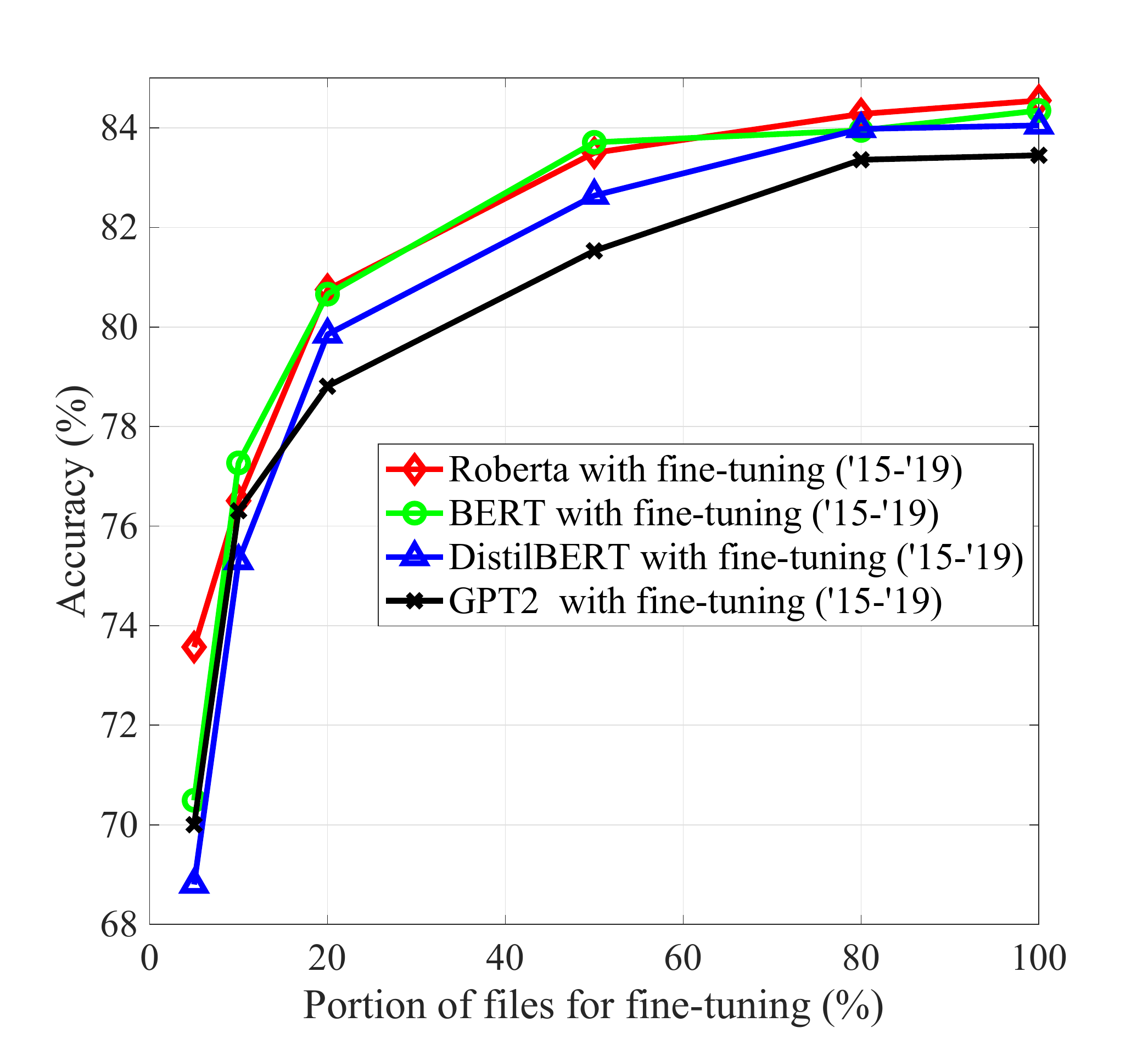}
\caption{Classification accuracy of different LLMs  vs. portion of 3GPP files used for fine-tuning.}
\label{fig:accuracy_vs_model}
\end{figure}

We start by comparing the performance of different LLMs fine-tuned with 3GPP files from 2015 to 2019 in terms of classification accuracy as illustrated in Fig. \ref{fig:accuracy_vs_model}. The selected models have the following sizes, BERT (117M), RoBERTa (125M), GPT-2 (124M) and DistilBERT (66M). Considering 100\% of the files, while it can be observed that all models experience relatively close accuracy, GPT-2 model encounters the weakest performance. This is motivated by the fact that for text classification, it is important for the large model to have concise and interpretable predictions features rather than generative capabilities, where the latter is the key element of GPT-2. Meanwhile, RoBERTa, the optimized version of BERT, demonstrates the strongest performance. It can be noticed further that the performance gap increases as the number of \ac{tdoc} files decreases. 

\begin{figure}[h!]
\centering
\includegraphics[width=1\linewidth]{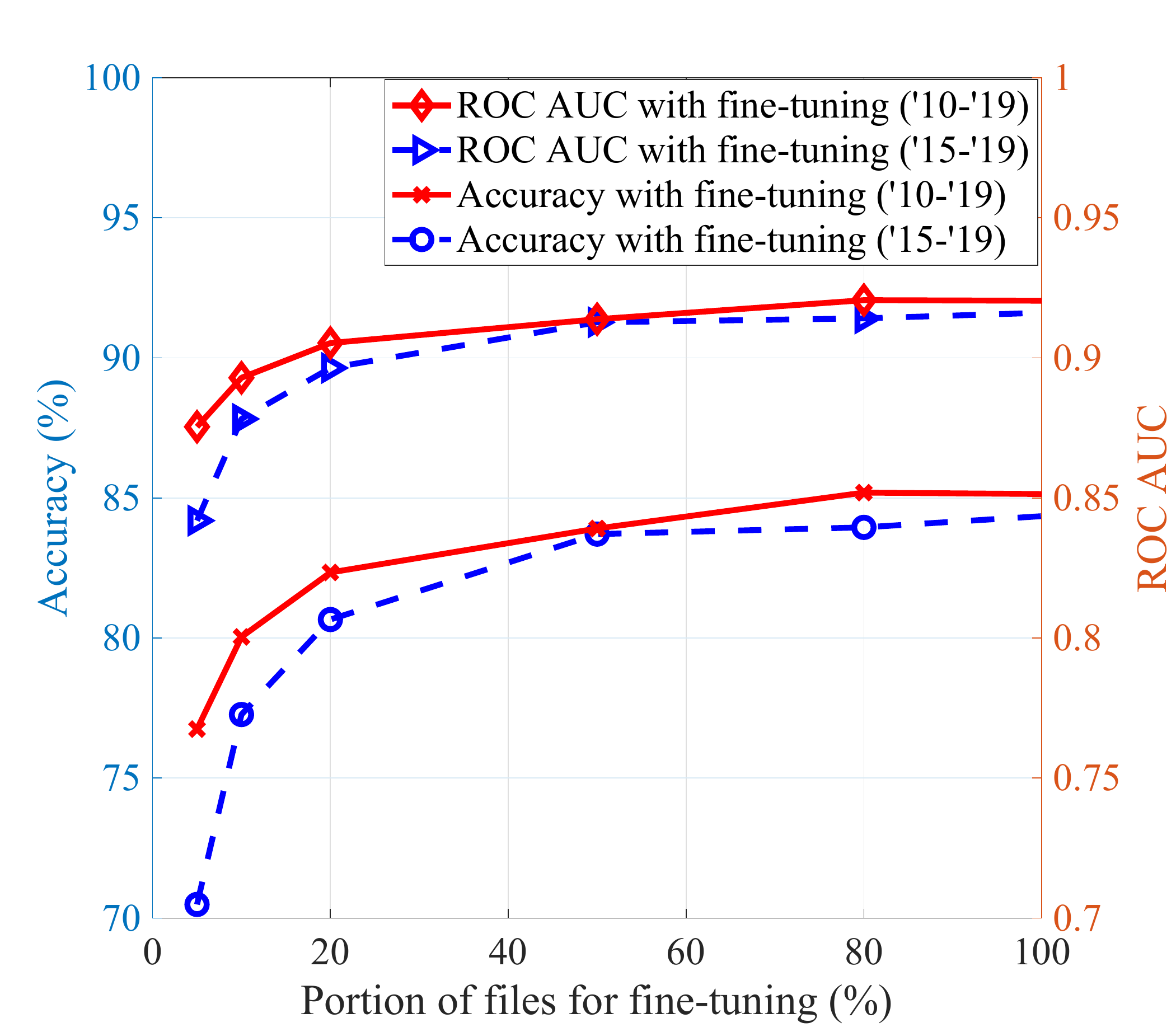}
\caption{Classificaiton accuracy and ROC-AUC vs. portion of 3GPP files used for fine-tuning BERT ('10-'19 and '15-'19). }
\label{fig:accuracy_vs_portion}
\end{figure}

In Fig. \ref{fig:accuracy_vs_portion}, we evaluate the accuracy of prediction and the receiver operating characteristic - area under the curve (ROC-AUC) as a function of the portion of textual segments used for fine-tuning a \ac{bert} model. We can observe that a \ac{bert} model fine-tuned to \ac{3gpp} files from 2015 to 2019 can achieve an accuracy performance around $80\%$ even if only $20\%$ of the text segments are used. Furthermore, although fine-tuning to Telecom language is essential, it can be demonstrated that \ac{tdoc} files from recent years are sufficient to provide the needed accuracy. On the other hand, when the number of files is relatively small (below $10\%$), data 2010-2015 produce better understanding of the Telecom technical language. 

The role of 3GPP \acp{wg} vary from one to another. Therefore the structure of files and especially the number of files available differs distinctly. For examples, RAN1 and RAN2 contains much more files than other \acp{wg} given that they are main categories in RAN (specifying PHY, MAC, RLC, PDCP layers), and hence, more activities pertinent to these layers are conducted within these two groups. To show the impact of different \acp{wg} on the performance of the classification, the accuracy of a BERT model fine-tuned by 3GPP files is illustrated in Table \ref{tab:accurary_vs_WGs} for different combinations of \acp{wg}. It can be noticed that the fine-tuned model can achieve  better classification accuracy for textual segment among RAN1, SA1 and CT1 than the combination of RAN1, RAN2 and RAN3. This is stemmed from the fact that \ac{tdoc} files belonging to different category number but fall within the same TSG are highly correlated, and therefore, the probability of error is higher. In contrast, technical files within different TSGs comprises relatively uncorrelated topics, and therefore, the model has a higher capability to distinguish between these different TSGs. The presented results in Table \ref{tab:accurary_vs_WGs} reveal that the test accuracy is determined mainly by the documents of RAN1, RAN2 and RAN3.
\begin{table}
\centering
\caption{Classification accuracy on different \ac{wg} combinations from BERT} \label{tab:accurary_vs_WGs}
\renewcommand{\arraystretch}{1.0}
\begin{tabular}{ | c | c | c | c | } 
\hline \hline
RAN & SA  & CT  & Accuracy ($\%$)    \\ 
\hline
1 &  1 & 1 & 98.05  \\
\hline
1,2,3 & None & None  & 88.90  \\ 
\hline
1,2,3 & 1 & 1  & 88.26  \\ 
\hline
1,2,3,4 & 2,5  &  None & 87.42  \\ 
\hline
1,2,3 & 1,2,3 & 1,3,4 & 86.61  \\ 
\hline
1,2,3,4 & 1,2,3,4 & 1,3,4,6  & 85.57  \\ 
\hline
1,2,3,4,5 & 1,2,3,4,5,6  &  1,3,4,6 & 84.35  \\ 
\hline\hline
\end{tabular}
\end{table}

\begin{figure}[h!]
\centering
\includegraphics[width=1\linewidth]{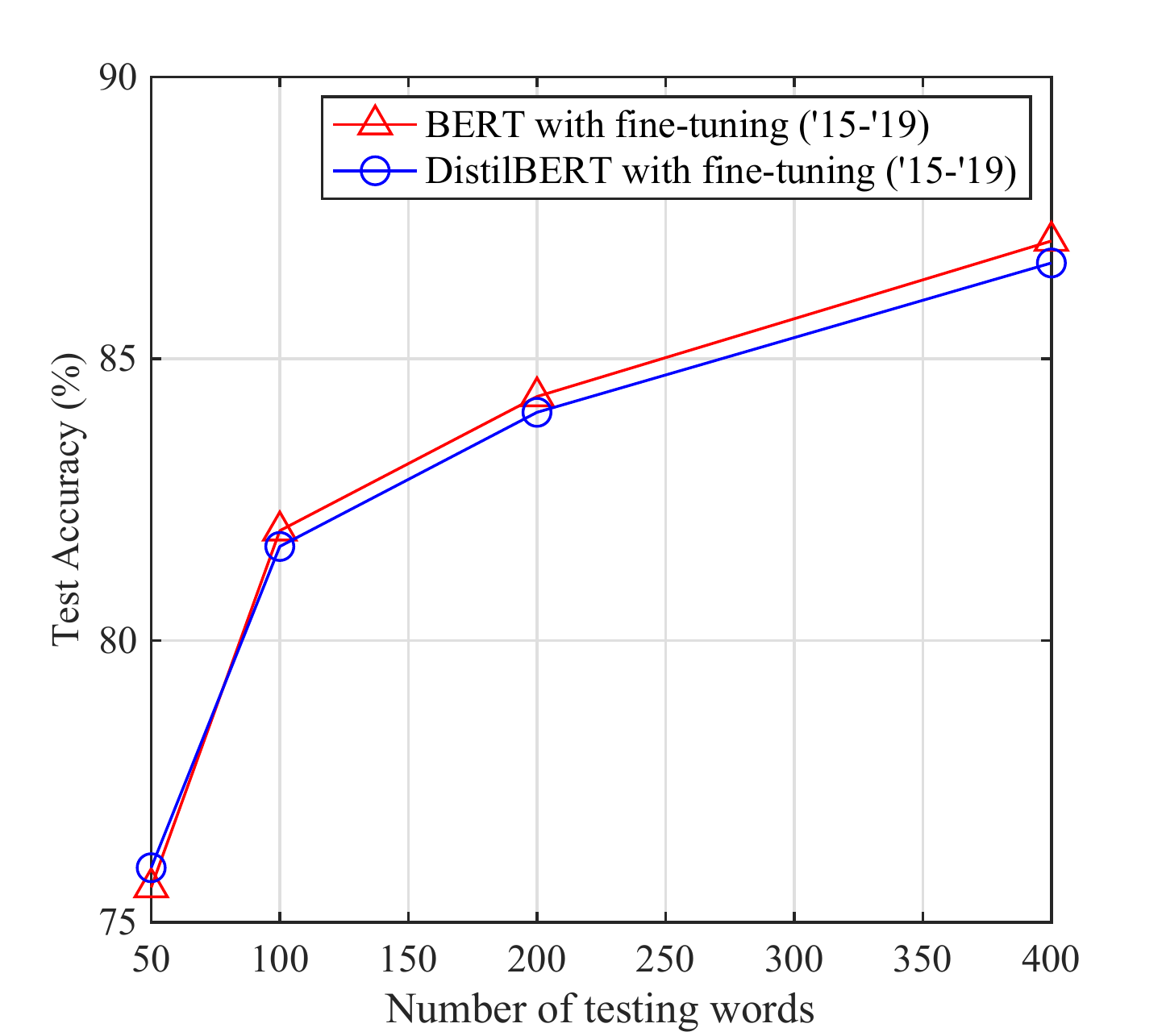}
\caption{Classification accuracy vs. maximum text segment length in words produced from fine-tuned BERT}
\label{fig:accuracy_vs_num_token}
\end{figure}

In Fig. \ref{fig:accuracy_vs_num_token} we evaluate the impact of the length of technical text segments to the accuracy of classification. This is a critical aspect to be studied, as it is important to know the minimum amount of text required by the tuned LLMs to realize accurate identification of the technical groups. We set the maximum number of words for training and validation to $200$ and we vary the number of words during the testing process. We can observe that the accuracy increases as the number of words grows. However, performance enhancement difference starts to decrease as well as the number of words increases, indicating the significant role of selecting the optimum size for a \ac{llm}, in order to strike a balance between performance and computing complexity. 

\section{Conclusion}
\label{conclusion}
Motivated by the promising potentials of \acp{llm}, in this paper, we proposed a framework for 3GPP technical documents identification, where we leveraged pre-trained language models, fine-tuned using 3GPP data, in order to allow the model to identify the 3GPP specification categories with the corresponding working group. In more details, we have considered \ac{bert}, DistilBERT, RoBERTa, and GPT-2 models, in which they are fine-tuned using 3GPP \acp{tdoc} belonging to TSGs, namely \ac{ran}, \ac{sa}, and \ac{ct}. The obtained results demonstrate the applicability of adapting a pre-trained language model into the Telecom domain, where all fine-tuned models showed accurate classification performance under different scenarios. It is important to emphasize the significance of developing \acp{llm} that are capable of understanding the Telecom language, as a cornerstone to enable autonomous networks driven by intelligent generative agents.  
\bibliographystyle{IEEEtran}
\bibliography{IEEEabrv,Refs}

\end{document}